\documentclass{article}

\usepackage{arxiv}

\usepackage[utf8]{inputenc}
\usepackage[T1]{fontenc}
\usepackage{lmodern}
\usepackage{mathptmx}              
\usepackage{microtype}
\usepackage{amsmath,amsfonts,amssymb}
\usepackage{nicefrac}
\usepackage{url}

\usepackage{booktabs}
\usepackage{tabularx}
\usepackage{multirow}
\usepackage{enumitem}
\usepackage[table]{xcolor}
\usepackage{graphicx}
\usepackage{tikz}
\usetikzlibrary{arrows.meta,positioning,calc,fit,backgrounds,shapes.misc}
\usepackage[ruled,vlined,linesnumbered]{algorithm2e}
\usepackage{listings}
\usepackage{caption}
\usepackage[most]{tcolorbox}
\usepackage[numbers,sort&compress]{natbib}
\usepackage[colorlinks=true,linkcolor=blue!50!black,citecolor=blue!50!black,
            urlcolor=blue!50!black]{hyperref}
\usepackage[capitalise,noabbrev]{cleveref}

\definecolor{ink}{HTML}{000000}
\definecolor{muted}{HTML}{555555}
\definecolor{hairline}{HTML}{CCCCCC}
\definecolor{accent}{HTML}{000000}
\definecolor{accentsoft}{HTML}{F4F4F5}
\definecolor{accent2}{HTML}{000000}
\definecolor{accent2soft}{HTML}{F4F4F5}
\definecolor{warn}{HTML}{000000}
\definecolor{warnsoft}{HTML}{F4F4F5}
\definecolor{llmfill}{HTML}{FFFFFF}
\definecolor{llmdraw}{HTML}{27272A}
\definecolor{detfill}{HTML}{F4F4F5}
\definecolor{detdraw}{HTML}{A1A1AA}
\definecolor{loopcol}{HTML}{18181B}
\definecolor{codebg}{HTML}{F7F7F7}
\definecolor{tablehead}{HTML}{F1F1F2}

\newcommand{\code}[1]{\texttt{#1}}
\newcommand{\sys}{ODR}
\newcommand{\sysfull}{Optimist Deep Research}

\setlist{itemsep=2pt,topsep=3pt,parsep=0pt}

\newtcolorbox{callout}[3][]{enhanced, breakable, colback=white,
  colframe=black!60, boxrule=0.4pt, arc=0pt, left=8pt, right=8pt,
  top=6pt, bottom=6pt, fonttitle=\bfseries\small, coltitle=black,
  attach title to upper, title={#3.\hspace{0.5em}}, after title={\par\smallskip}, #1}

\SetKwComment{Comment}{$\triangleright$\ }{}
\SetAlFnt{\small}

\title{Optimist Deep Research: An Iterative, State-Centric Agent for Autonomous Literature Research}

\author{
  Gihan Panapitiya \\
}

\date{October 2026}

\renewcommand{\headeright}{Technical Report}
\renewcommand{\undertitle}{Technical Report}
\renewcommand{\shorttitle}{Optimist Deep Research}

\hypersetup{
  pdftitle={Optimist Deep Research: An Iterative, State-Centric Agent for Autonomous Literature Research},
  pdfauthor={Gihan Panapitiya},
  pdfkeywords={LLM agents, literature review automation, retrieval-augmented generation, citation snowballing, evidence synthesis, LangGraph},
}

\begin{document}
\maketitle

\begin{abstract}
Most retrieval-augmented systems treat literature search as a single retrieval step followed by summarization. Real literature research works differently: the investigator's understanding of the question changes as they read, new subquestions appear, and evidence is weighed, not just collected. We present Optimist Deep Research, an open-source agent that runs literature research as an \emph{iterative evidence-acquisition process} over an explicit, mutable research state. A research question is broken into a tree of subquestions, each served by \emph{search intents} that are translated into source-specific queries over arXiv, OpenAlex, Crossref, PubMed, Semantic Scholar and the web. Results are merged into a canonical, provenance-preserving document registry through a conservative four-tier deduplication hierarchy. The registry is expanded by citation snowballing until a saturation criterion is met, and then screened, ranked and allocated to subquestions round-robin. Documents are read in parallel into a structured \emph{evidence store} of claims with support, location, type and strength. A critic reviews the state along eight dimensions and returns \emph{research actions} rather than a binary verdict: it can add, split, rephrase, deprioritize or close subquestions, so the research plan is an output of the investigation as well as an input. The final report is written section by section from the evidence store, and every citation resolves to a retrieved document. We describe the design, the deterministic machinery that keeps a language-model pipeline auditable, and a case study that read 28 of 237 retrieved documents in about three and a half minutes. We also report weaknesses the case study exposed.
\end{abstract}


\section{Introduction}

Large language models (LLMs) combined with retrieval have become a standard way to answer knowledge-intensive questions~\citep{lewis2020rag}. For \emph{literature research}, meaning the task of establishing what a body of scholarly work says about a question, a single round of retrieve-then-summarize falls short in three ways.

First, the right decomposition of a research question is rarely clear at the start. It becomes clear while reading: a paper introduces terminology the investigator did not know, shows that a subquestion has two parts that need separate evidence, or shows that another subquestion is a dead end. Second, the output of reading should be \emph{evidence}, not summaries. A summary turns a paper into a single opinion. Evidence keeps each claim separate from what supports it, so a later stage can ask whether a claim is established by several independent studies or asserted by one. Third, systematic reviewers have long known that keyword search alone misses relevant work, and that following citations backward and forward is needed for coverage~\citep{wohlin2014guidelines,kitchenham2004procedures}.

\sysfull{} (\sys{})\footnote{Code: \url{https://github.com/gihanpanapitiya/optimist_research}.} is built around these observations. Its central design principle is
that \textbf{the research plan is not fixed}: subquestions, intents,
documents, evidence, contradictions and coverage estimates all live in one
explicit state, and every stage of the pipeline reads and changes that
state. The contributions of this paper are:

\begin{callout}[unbreakable]{accent}{Contributions}
\begin{enumerate}[leftmargin=*]
  \item A \textbf{state-centric architecture} for literature research, in
    which a five-level hierarchy (question $\to$ subquestion $\to$ intent
    $\to$ query $\to$ source) is first-class, mutable data. Its reducers
    remain correct under parallel fan-out (\cref{sec:state}).
  \item A \textbf{conservative canonicalization hierarchy} for scholarly
    records gathered from heterogeneous sources. It uses numeral vetoes,
    Unicode folding and verified identifier lookups, so false merges, which
    silently destroy evidence, are rare (\cref{sec:canon}).
  \item \textbf{Deterministic ranking, allocation and saturation rules} that
    limit LLM calls to the points where judgement is needed. These include a
    round-robin reading allocator that prevents one subquestion from using
    up the reading budget, and a direction-aware snowballing policy
    (\cref{sec:rank,sec:snowball}).
  \item A \textbf{structured evidence store} and a \textbf{critic that
    returns plan edits}, which together close an adaptive research loop
    whose termination is controlled by both quality and budget
    (\cref{sec:evidence,sec:critic}).
  \item An \textbf{end-to-end case study} that reports what the system did,
    including several failure modes found in the process
    (\cref{sec:case}).
\end{enumerate}
\end{callout}

\section{Background and Related Work}
\label{sec:related}

\paragraph{Retrieval-augmented and agentic LLMs.}
Retrieval augmentation conditions generation on retrieved
passages~\citep{lewis2020rag}. Agentic variants interleave reasoning with
tool use~\citep{yao2023react} or with retrieval at each reasoning
step~\citep{trivedi2023ircot,wei2022cot}, and decomposition methods split a
question into simpler subquestions that are answered in
turn~\citep{press2023selfask,zhou2023leasttomost}. Self-RAG~\citep{asai2024selfrag}
lets the model decide when to retrieve and critique its own retrieved
passages and generations, and more recent work trains reasoning models to
issue searches during long reasoning chains, either through prompting
frameworks~\citep{li2025searcho1} or reinforcement learning in live web
environments~\citep{zheng2025deepresearcher}.
Self-reflection methods have the model critique and revise its own
output~\citep{shinn2023reflexion,madaan2023selfrefine}. In all of these, the
decomposition and the critique live inside the model's context. \sys{}
moves both into explicit state. The subquestion tree is data that every
stage reads, and the critic's output is a set of typed edits to that tree
plus new search intents, which are applied deterministically.

\paragraph{Deep research agents and their evaluation.}
Commercial ``deep research'' products~\citep{openai2025deepresearch} popularized agents that browse for many minutes and return long, cited reports. STORM~\citep{shao2024storm} outlines Wikipedia-style articles through perspective-guided question asking, and Co-STORM~\citep{jiang2024costorm} adds a human to the conversation. A growing set of benchmarks measures these agents: BrowseComp for hard-to-find facts~\citep{wei2025browsecomp}, DeepResearch Bench~\citep{du2025deepresearchbench}, ResearchRubrics~\citep{sharma2025researchrubrics} and DEER~\citep{han2025deer} for report quality against expert criteria, Total Recall QA for verifiable recall~\citep{rafiee2026totalrecall}, and IDRBench for the value of clarifying interaction~\citep{feng2026idrbench}. Two findings bear directly on our design. First, SAGE~\citep{hu2026sage} finds that current agents struggle with reasoning-intensive scientific retrieval, and that BM25 outperforms LLM-based retrievers inside an agent because the agents issue keyword-oriented sub-queries. \sys{} accepts that sources differ and routes each search intent into source-specific queries rather than one
query string for all of them. Second, these benchmarks evaluate an
end-to-end answer. \sys{}'s state records why each document was found and
read, which makes stage-level evaluation possible (\cref{sec:discussion}).

\paragraph{LLM systems for scholarly search and synthesis.}
PaperQA~\citep{lala2023paperqa} answers scientific questions by retrieving
and reading full-text papers. OpenScholar~\citep{asai2024openscholar} and
Ai2 Scholar QA~\citep{singh2025scholarqa} build open retrieval pipelines
over scholarly corpora for citation-backed synthesis. For paper search
specifically, PaSa~\citep{he2025pasa} trains an agent with reinforcement
learning to invoke search tools, read papers and expand through their
references, and SPAR~\citep{shi2025spar} uses multiple agents with
reference-chain query decomposition and query evolution. LitSearch~\citep{ajith2024litsearch}
provides a retrieval benchmark for literature search, and
LitLLMs~\citep{agarwal2024litllms} studies literature-review writing as
keyword extraction, retrieval, re-ranking and generation. Knowledge
Navigator~\citep{katz2024knowledgenavigator} organizes the results of a
broad query into a topic hierarchy that a user refines interactively, and
AutoSurvey~\citep{wang2024autosurvey} drafts full surveys from retrieved
papers via outlines and section-wise generation. Further along, Agent
Laboratory~\citep{schmidgall2025agentlab} and The AI
Scientist~\citep{lu2024aiscientist} place literature review as one stage in
an automated research pipeline. Most of these systems either fix the
decomposition at the start or adapt only the next query. \sys{} makes the
plan itself revisable: the critic can add, split, rephrase, deprioritize or
close subquestions as the evidence comes in. It also keeps an explicit
evidence store of claims, rather than passing retrieved text straight to a
writer. The closest systems engineering is AgentR~\citep{samanta2026agentr},
which persists the intents, queries, paper assessments and gap analyses of
an LLM literature-review workflow as durable artifacts for recovery and
auditing. \sys{} shares that view of state, but uses it to drive adaptive
planning rather than mainly for fault tolerance.

\paragraph{Citation grounding.}
Work on citation-grounded generation~\citep{gao2023alce} shows that LLMs
often cite sources that do not support the claim. The problem persists in
deep research agents: across ten models and agents, 3--13\% of citation
URLs are hallucinated and 5--18\% do not resolve, with deep research agents
hallucinating at higher rates than search-augmented
LLMs~\citep{rao2026referencehallucinations}, and cited content is often
accessible and relevant but not factually
supportive~\citep{onweller2026citedbutnotverified}. These studies detect
bad citations after the fact. \sys{} prevents the first kind
structurally. Citations are document identifiers taken from the evidence
passed to each report section, unresolvable identifiers are removed, and
the reference list is built from the document registry rather than
generated by the model. Whether a cited claim is actually supported still
depends on the reader's extraction, which is why the evidence store
records support and strength separately from the claim.

\paragraph{Systematic review methodology and its automation.}
Kitchenham's procedures~\citep{kitchenham2004procedures, kitchenham2007guidelines} and the PRISMA
statement~\citep{page2021prisma} formalize search, screening, eligibility
and synthesis for human-conducted reviews. Wohlin~\citep{wohlin2014guidelines}
describes backward and forward snowballing as an iterative procedure that
runs until no new relevant papers are found, and tool support for forward
snowballing remains uneven~\citep{januario2025forwardsnowballing}. Surveys
of AI for systematic reviews~\citep{bolanos2024aireviews} and a scoping
review of LLM use~\citep{lieberum2025llmsr} find LLMs applied most often to
search, study selection and data extraction, but judge them not yet ready
for unsupervised use. Individual studies evaluate LLMs for Boolean query
formulation~\citep{wang2023booleanquery} and title--abstract
screening~\citep{syriani2023screening}. InsightAgent~\citep{qiu2025insightagent}
keeps a human in the loop and reports completing a systematic review in
hours. SciLitBench~\citep{zabaleta2026scilitbench} evaluates screening and
extraction as linked stages, and ARISMA~\citep{moghaddam2026arisma}
proposes reporting guidelines for AI-assisted reviews. \sys{} implements
an automated, budgeted version of the search--screen--snowball cycle. Its
saturation criterion directly operationalizes ``no new papers,'' and its
provenance records are the kind of audit trail such guidelines ask for.
It does not claim to replace a protocol-driven systematic review.

\paragraph{Record linkage.}
Deduplicating bibliographic records is an instance of record
linkage~\citep{fellegi1969record,elmagarmid2007duplicate}. Our hierarchy
puts persistent identifiers first and fuzzy matching last, and it is tuned
for precision over recall. The reason is specific to this setting: a missed
duplicate costs one redundant read, while a false merge silently attributes
one paper's evidence to another.

\section{Design Principles}
\label{sec:principles}

Five principles shaped the implementation.

\begin{description}[leftmargin=1.2em,style=nextline]
  \item[P1. Research is a state, not a pipeline.] The state model, rather
    than the graph topology, encodes the method. Every node is a function
    from state to a \emph{delta} on state.
  \item[P2. The plan is mutable.] Subquestions carry status, priority,
    coverage and origin (\code{planner}, \code{critic} or
    \code{discovery}), and the critic can rewrite them.
  \item[P3. Evidence, not summaries.] The unit of output from reading is a
    claim paired with its support, location, type and strength.
  \item[P4. Use LLMs only for judgement.] Ranking, reading allocation,
    snowball seed selection, saturation, deduplication, termination and
    reference assembly are deterministic code, so they are testable and
    auditable.
  \item[P5. Budgets are limits, not targets.] The number of documents
    retrieved is never a stopping criterion. Budgets cap cost, and the
    critic decides sufficiency.
\end{description}

\section{Architecture}
\label{sec:arch}

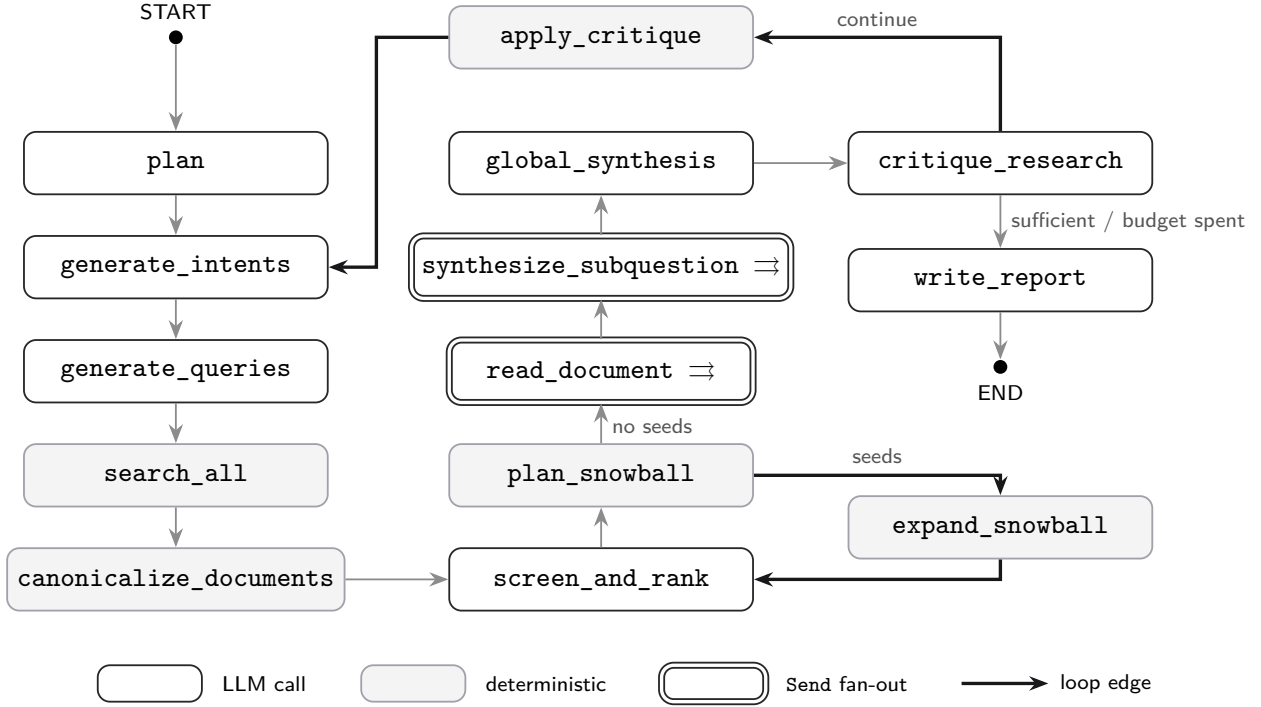
\begin{figure}[t]
\centering
\resizebox{\textwidth}{!}{%
\begin{tikzpicture}[
    font=\small\sffamily,
    >={Stealth[length=2.2mm]},
    box/.style={rounded corners=4pt, minimum width=3.5cm, minimum height=0.72cm,
                align=center, line width=0.6pt, font=\ttfamily\footnotesize},
    llm/.style={box, fill=llmfill, draw=llmdraw},
    det/.style={box, fill=detfill, draw=detdraw},
    fan/.style={double, double distance=1.1pt},
    term/.style={circle, fill=black, inner sep=1.6pt},
    lbl/.style={font=\scriptsize\sffamily, text=muted},
    flow/.style={->, line width=0.6pt, draw=black!45},
    loop/.style={->, line width=1.1pt, draw=loopcol},
  ]
  \node[term, label={[font=\scriptsize\sffamily]above:START}] (start) at (0,1.45) {};
  \node[llm] (plan)    at (0, 0.00) {plan};
  \node[llm] (intents) at (0,-1.20) {generate\_intents};
  \node[llm] (queries) at (0,-2.40) {generate\_queries};
  \node[det] (search)  at (0,-3.60) {search\_all};
  \node[det] (canon)   at (0,-4.80) {canonicalize\_documents};

  \node[llm] (screen)  at (4.9,-4.80) {screen\_and\_rank};
  \node[det] (snowp)   at (4.9,-3.60) {plan\_snowball};
  \node[llm,fan] (read)  at (4.9,-2.40) {read\_document $\rightrightarrows$};
  \node[llm,fan] (sqsyn) at (4.9,-1.20) {synthesize\_subquestion $\rightrightarrows$};
  \node[llm] (gsyn)    at (4.9, 0.00) {global\_synthesis};

  \node[det] (expand)  at (9.5,-4.20) {expand\_snowball};
  \node[llm] (critic)  at (9.5, 0.00) {critique\_research};
  \node[det] (apply)   at (4.9, 1.45) {apply\_critique};
  \node[llm] (report)  at (9.5,-1.35) {write\_report};
  \node[term, label={[font=\scriptsize\sffamily]below:END}] (end) at (9.5,-2.35) {};

  \draw[flow] (start) -- (plan);
  \draw[flow] (plan) -- (intents);
  \draw[flow] (intents) -- (queries);
  \draw[flow] (queries) -- (search);
  \draw[flow] (search) -- (canon);
  \draw[flow] (canon) -- (screen);
  \draw[flow] (screen) -- (snowp);
  \draw[flow] (snowp) -- node[lbl, right, pos=0.45] {no seeds} (read);
  \draw[flow] (read) -- (sqsyn);
  \draw[flow] (sqsyn) -- (gsyn);
  \draw[flow] (gsyn) -- (critic);
  \draw[flow] (critic) -- node[lbl, right] {sufficient / budget spent} (report);
  \draw[flow] (report) -- (end);

  \draw[loop] (snowp.east) -| node[lbl, above, pos=0.25] {seeds} (expand.north);
  \draw[loop] (expand.south) |- (screen.east);

  \draw[loop] (critic.north) |- node[lbl, above, pos=0.75] {continue} (apply.east);
  \draw[loop] (apply.west) -| ($(intents.east)+(0.55,0)$) -- (intents.east);

  \node[llm, minimum width=1.2cm, minimum height=0.42cm, font=\scriptsize\sffamily] (l1) at (-0.3,-6.0) {};
  \node[right=1mm of l1, font=\scriptsize\sffamily] (t1) {LLM call};
  \node[det, minimum width=1.2cm, minimum height=0.42cm, font=\scriptsize\sffamily, right=5mm of t1] (l2) {};
  \node[right=1mm of l2, font=\scriptsize\sffamily] (t2) {deterministic};
  \node[llm, fan, minimum width=1.2cm, minimum height=0.42cm, font=\scriptsize\sffamily, right=5mm of t2] (l3) {};
  \node[right=1mm of l3, font=\scriptsize\sffamily] (t3) {\texttt{Send} fan-out};
  \draw[loop] ($(t3.east)+(0.5,0)$) -- ++(1.0,0) node[right, font=\scriptsize\sffamily, text=ink] {loop edge};
\end{tikzpicture}}
\caption{The research graph. White nodes call an LLM and grey-filled nodes are
deterministic. Heavy black edges are loops. Double-bordered nodes fan out with LangGraph's \texttt{Send},
one branch per document or subquestion. The inner loop (right, lower)
expands the citation graph and re-screens. The outer loop (top) applies the
critic's research actions and re-enters at intent generation, not at
planning, so the existing plan is edited rather than regenerated.}
\label{fig:graph}
\end{figure}

\sys{} is implemented as a LangGraph~\citep{langgraph} state graph
(\cref{fig:graph}) built around two loops. The \textbf{inner loop} widens
the corpus: it follows the citations of relevant documents already found,
screens the new candidates, and stops once further rounds turn up few new
documents (\cref{sec:snowball}). The \textbf{outer loop} steers the
research: after each round of reading, a critic reviews
the evidence, edits the research plan, and sends the system back to search
for what is still missing (\cref{sec:critic}).

\subsection{The research state}
\label{sec:state}

The state records the investigation as a hierarchy:
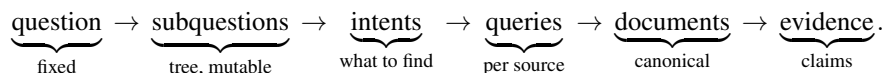
\[
\underbrace{\text{question}}_{\text{fixed}} \;\to\;
\underbrace{\text{subquestions}}_{\text{tree, mutable}} \;\to\;
\underbrace{\text{intents}}_{\text{what to find}} \;\to\;
\underbrace{\text{queries}}_{\text{per source}} \;\to\;
\underbrace{\text{documents}}_{\text{canonical}} \;\to\;
\underbrace{\text{evidence}}_{\text{claims}} .
\]

Subquestions and intents are kept as separate levels on purpose. A \emph{subquestion} states what we need to \emph{know}, for example ``How does episodic memory compare with semantic memory for long-horizon tasks?'' A \emph{search intent} describes a kind of document that would help answer it, for example ``empirical comparisons of episodic versus semantic memory on long-horizon benchmarks.'' A subquestion can have several intents. This separation lets the system change \emph{how} it searches (adding, closing or rewording intents and their queries) without changing \emph{what} it is trying to find out. Each intent also records its \emph{yield}: how many documents its queries found, and how many of those were screened as relevant. The critic uses these yields to judge whether an intent is still productive.

\paragraph{Reducers under fan-out.}
Parallel graph branches write to shared channels, so each channel has a
merge function (a \emph{reducer}) that must be associative and must
tolerate two branches having seen the same baseline. Three rules make this
work.

\begin{itemize}[leftmargin=*]
  \item \textbf{Budgets are written as deltas.} Each reader writes
    \code{documents\_read=1} and the reducer sums them, so twenty parallel
    readers advance the counter by exactly twenty. If branches wrote
    running totals, each one would need to know what the others had done.
  \item \textbf{Documents merge field by field.} The more informative value
    wins: the longer abstract, the union of sources and provenance, the
    maximum of each subquestion's relevance score, and the minimum
    snowball depth. A sparse web hit therefore never degrades a rich
    record.
  \item \textbf{Document status never moves backward.} Several branches
    can return copies of the same document, and some copies are out of
    date. For example, a reader may report a document as \code{read}
    while another branch, working from an earlier copy, still reports it
    as \code{selected}. To merge these, statuses are ranked
    \[
      \texttt{candidate} < \texttt{selected}
      < \texttt{failed} < \texttt{read},
    \]
    and the higher one is kept. A finished read is therefore never
    undone, and a document whose read failed stays \code{failed} instead
    of going back into the reading queue in every later iteration.
\end{itemize}

Per-iteration work lists (query plans, pending hits, the reading queue)
replace their previous contents rather than accumulating, so an iteration
starts from a fresh query set. Evidence and contradictions are append-only,
with duplicates removed by content hash.

\subsection{Model routing}

Each LLM role (planner, intent, query, screener, reader, synthesizer,
critic, writer) is routed independently to any \code{provider:model}
specification. The roles differ in cost by orders of magnitude. Screening
and reading run once per \emph{document} (or batch of documents), while planning
and critique run once per \emph{iteration}. Per-role routing is therefore
the main cost control: a capable model on the judgement roles and a cheaper
one on the high-volume roles. All structured calls go through one helper
that validates output against a Pydantic schema and returns \code{None} on
failure. Each node then uses an explicit fallback, so a malformed response
degrades a single step instead of stopping the run.

\section{Planning and Retrieval}
\label{sec:retrieval}

\paragraph{Decomposition.}
The planner is instructed to produce the \emph{minimum} set of meaningful
subquestions, each with a rationale and a priority. Because each intent's
yield is recorded, the value of decomposition can be measured: setting
\code{max\_subquestions}~$=1$ gives a single-query baseline.

\paragraph{Source routing.}
A single query string cannot serve every scholarly API. arXiv uses literal
keyword matching with terms combined by AND, Semantic Scholar ranks
natural-language queries semantically, and PubMed rewards controlled
vocabulary. \sys{} therefore first \emph{selects} the sources relevant to
an intent and then generates a query specialized for each, up to
$\min(4, |\text{sources}|)$ queries per intent. Each (source, query) pair
counts as one search operation against the budget. Intents from
higher-priority subquestions are planned first, so if the budget runs out,
the least important work is the work that is lost. Intents that would
produce no new queries are closed.

\paragraph{Sources.}
\Cref{tab:sources} lists the supported sources. arXiv, OpenAlex, Crossref
and PubMed need no key. Semantic Scholar has three settings (\code{auto},
\code{on}, \code{off}), and under \code{auto} it is used only when an API
key is present. Its unauthenticated pool is shared per IP address and
returns HTTP~429 even under light load, so leaving it on without a key
spends search budget for nothing. HTTP search runs concurrently inside a
single node with a semaphore, rather than as one graph branch per query.
Each source already enforces its own rate limit (arXiv asks for one request
every three seconds), and per-query branches would bloat the state without
any benefit.

\begin{table}[t]
\centering
\small
\caption{Literature sources. ``Citation graph'' is the direction of
snowballing that the source supports.}
\label{tab:sources}
\begin{tabular}{@{}llll@{}}
\toprule
Source & Key & Citation graph & Best for \\
\midrule
arXiv            & ---           & none        & very recent CS/ML preprints \\
OpenAlex~\citep{priem2022openalex} & --- & both & broad coverage, DOIs, snowballing \\
Crossref         & ---           & backward    & deposited reference lists \\
PubMed           & optional      & none        & biomedical literature \\
Semantic Scholar~\citep{kinney2025semanticscholaropendata} & effectively required & both & richest citation graph \\
Web (Tavily)     & required      & none        & grey literature, documentation \\
\bottomrule
\end{tabular}
\end{table}

\section{Canonicalization}
\label{sec:canon}

A single paper often arrives from several sources: arXiv as a preprint,
Crossref with the published DOI, OpenAlex with a citation count, and the
web as a landing page. These records are merged into one canonical document
\emph{before} reading, which is the expensive stage, and every
\code{Provenance} record (subquestion, intent, source, query, rank, and
whether the document was found by search or by snowballing) is kept. This
means a paper retrieved for one subquestion can serve another without being
retrieved again, and retrieval coverage can be evaluated afterwards.

\paragraph{Matching hierarchy.} Candidates are matched in order of
decreasing strength:
\begin{enumerate}[leftmargin=*]
  \item DOI (normalized);
  \item a persistent identifier: arXiv (version suffix stripped), PubMed,
    OpenAlex or Semantic Scholar;
  \item normalized title plus first-author surname;
  \item title similarity of at least $0.93$ (\code{SequenceMatcher}
    ratio) together with a corroborating signal (year or authors).
\end{enumerate}
Tier~4 is the only fuzzy tier, and it is intentionally strict. The
threshold of $0.93$ is high because papers in a series (``\ldots{} Part
II'', ``\emph{Method} v2'') score between $0.85$ and $0.92$ and must not be
merged.

\paragraph{Numeral veto.} Numbers in a title almost always distinguish
separate works: \emph{Llama~2} and \emph{Llama~3}, \emph{GPT-3} and
\emph{GPT-4}, \emph{Part~II} and \emph{Part~III}. These pairs differ by one
or two characters and would pass any similarity threshold. We therefore
compute a \emph{numeric signature} (the sorted multiset of Arabic and Roman
numerals in the title), and any mismatch vetoes a fuzzy merge.

\paragraph{Truncation.} A record may carry a short title or drop a
subtitle. A prefix match is accepted only when the shorter title has at
least three distinctive tokens, because no length threshold separates a
specific stub (``llm planner'') from a generic one (``deep learning'').
Containment, meaning all distinctive tokens of the shorter title appear in
the longer one, requires at least four.

\paragraph{Unicode folding.} Names and titles are NFKD-normalized, combining
marks are removed, and dash variants are unified before comparison. This
fixed a real failure: OpenAlex writes \emph{Romera‐Paredes} with
U+2010~HYPHEN, which compared as ``paredes'' against ``romera-paredes'' and
caused a correct match to be rejected. A full title match is also
\emph{never} vetoed because the author lists do not overlap. Sources
render names too inconsistently (initials only, collective authors,
transliterations) for that signal to override a matching title. Author
agreement is required only when the title evidence is weak.

\paragraph{Verified identifier lookups.} Some arXiv DOIs in OpenAlex are
attached to the wrong record. Looking up \code{10.48550/arXiv.2201.11903}
(Chain-of-Thought prompting~\citep{wei2022cot}) returns an unrelated paper.
Snowballing from that record would add someone else's references to the
corpus under the seed's name. Every OpenAlex identifier lookup is therefore
checked against the document's title and authors, with a
punctuation-stripped title search as a fallback. A lookup that cannot be
verified is rejected and logged.

\section{Screening, Ranking and Reading Allocation}
\label{sec:rank}

\paragraph{Screening.} An LLM scores batches of documents (ten by default)
against every active subquestion and returns a relevance score
$r_q(d)\in[0,1]$ and a one-line note for each document. The prompt tells
the model to judge substance rather than keyword overlap, since shared
terminology is the most common cause of a bad screening decision. Scores
are merged by maximum. Subquestion identifiers returned by the model are
\emph{repaired} before matching. Some models echo the whole prompt line
(``\code{RQ1: How do agents\ldots}'') where the bare id \code{RQ1} is
expected, and discarding those entries would score every document zero.
When a new subquestion is added, the existing corpus is screened only
against it, so document--subquestion pairs that already have scores are
not paid for twice.

\paragraph{Ranking.} The rank score combines relevance with the retrieval
objectives from the methodology:
\begin{equation}
\label{eq:rank}
s(d) = \lambda(d)\Big[
  w_r \max_q r_q(d) + w_c C(d) + w_t T(d) + w_s Q(d)
  + w_m \min\!\big(\tfrac{|S(d)|}{3},1\big) + w_p P(d) + w_n N(d)
\Big],
\end{equation}
where $S(d)$ is the set of sources that returned $d$, $P(d)\in\{0,1\}$ is a
heuristic for primary research, and $\lambda(d)=0.6$ when $d$ has no
abstract in abstract-only mode (1 otherwise). The default weights are
$w_r=1.0$, $w_c=0.15$, $w_t=0.20$, $w_s=0.10$, $w_m=0.10$, $w_p=0.10$ and
$w_n=0.15$. Relevance dominates by design, and the other terms act as
tie-breakers. Citation impact is normalized by the paper's age:
\begin{equation}
C(d) = \min\!\left(\frac{\log(1 + c_d / a_d)}{\log 61},\, 1\right),
\qquad a_d = \max(1,\; y_{\text{now}} - y_d + 1),
\end{equation}
so a paper with sixty citations per year saturates the term. Raw citation
counts would push down work from the last year or two, which is exactly the
work a question about recent developments needs. A missing citation count
scores $0.3$ rather than zero, since missing data is not evidence of
missing impact. Recency $T(d)$ is $1$ for papers at most one year old and
falls linearly to $0$ at fifteen years. Source quality $Q(d)$ is $1.0$ for
a DOI with a venue, $0.85$ for other indexed records, $0.7$ for preprints
without a DOI, and $0.35$ for web-only records. Novelty is one minus the
largest title-token Jaccard similarity to any document already read,
\begin{equation}
N(d) = 1 - \max_{d' \in \mathcal{R}} J\big(\tau(d), \tau(d')\big),
\end{equation}
which keeps the reading queue from filling with near-duplicates once an
area is well covered.

\paragraph{Reading allocation.} A single global top-$N$ cut lets one
well-served subquestion take the whole reading budget and leave nothing
for the others, which works against coverage. \sys{} instead allocates
reads round-robin across subquestions in priority order
(\cref{alg:select}). A second phase then attaches every other subquestion
that a selected document is relevant to. Reading one document against
several subquestions is a single LLM call, so these attachments cost no
reading budget. This is how a paper retrieved for RQ1 also comes to serve
RQ3. The queue is sorted by rank score, so if the budget runs out partway
through, the least valuable reads are the ones lost.

\begin{algorithm}[t]
\caption{Round-robin reading allocation}
\label{alg:select}
\KwIn{ranked corpus $\mathcal{D}$; active subquestions $\mathcal{Q}$ sorted by priority;
      threshold $\theta$; iteration cap $K$; per-subquestion cap $k$}
\KwOut{assignments $A: d \mapsto$ set of subquestions}
\ForEach{$q \in \mathcal{Q}$}{
  $L_q \leftarrow [\,d \in \mathcal{D} : r_q(d)\ge\theta,\ q \notin \mathrm{readFor}(d)\,]$
  sorted by $0.6\,r_q(d) + 0.4\,s(d)$ descending\;
}
$\mathit{selected}\leftarrow\emptyset$\;
\Repeat(\Comment*[f]{phase 1: costs budget}){$|\mathit{selected}|\ge K$ or no subquestion advanced}{
  \ForEach{$q \in \mathcal{Q}$ with $|\{d: q\in A(d)\}| < k$}{
    take next $d\in L_q$ with $q\notin A(d)$; \ $A(d)\leftarrow A(d)\cup\{q\}$;
    \ $\mathit{selected}\leftarrow \mathit{selected}\cup\{d\}$\;
  }
}
\ForEach(\Comment*[f]{phase 2: free attachments}){$q\in\mathcal{Q}$}{
  \ForEach{$d\in L_q \cap \mathit{selected}$ while $|\{d: q\in A(d)\}| < k$}{
    $A(d)\leftarrow A(d)\cup\{q\}$\;
  }
}
\Return{$A$, ordered by $s(d)$ descending}\;
\end{algorithm}

\section{Citation Snowballing}
\label{sec:snowball}

Snowballing runs after ranking and starts only from documents that screened
as clearly relevant ($r_q\ge 0.7$) and were found by direct search. Seeds
are chosen without an LLM call, because the shape of the corpus already
indicates which \emph{direction} is useful:

\begin{itemize}[leftmargin=*]
  \item If a subquestion's relevant evidence is clustered in recent years
    (a span of at most 3 years with median year $\ge 2022$), its
    foundations are missing. The system expands \textbf{backward} from the
    most relevant recent paper, whose reference list is a curated path into
    the area's earlier work.
  \item If the median year is old ($\le 2020$), or the corpus is mixed,
    recent developments are missing. The system expands \textbf{forward}
    from the most-cited paper, which has the most citing work to find.
\end{itemize}

Seeds explicitly requested by the critic take precedence over automatic
ones. Each source supports a different chain of directions:
\[
\text{backward: } \text{S2}^{\dagger} \to \text{OpenAlex} \to \text{Crossref},
\qquad
\text{forward: } \text{S2}^{\dagger} \to \text{OpenAlex},
\]
where $\dagger$ marks Semantic Scholar, used only with a key. OpenAlex
comes first because each edge it returns is a full record with an
abstract, so the candidate can be read. Crossref often returns
\emph{more} backward edges (43 against 22 for one IEEE paper), but roughly
half lack a DOI. Its DOI-bearing entries are therefore looked up in
OpenAlex in batches of 50 to recover titles, authors and abstracts.

\paragraph{Saturation.} Following Wohlin's stopping rule (``until no new
papers are found'')~\citep{wohlin2014guidelines}, round $k$ is assigned a
yield
\begin{equation}
\rho_k = \frac{\#\,\text{new canonical documents in round } k}
              {\#\,\text{candidates examined in round } k},
\end{equation}
measured in \emph{new canonical documents} rather than raw hits. A round
that returns 200 references already in the corpus has saturated, even
though it returned plenty of data. Snowballing stops when two
\emph{consecutive} rounds satisfy $\rho < \rho_{\min}$ (default
$\rho_{\min}=0.10$), when the round budget is used up, or when the reading
budget is used up. Requiring two rounds keeps a single seed with poor
citation coverage from ending exploration too early. Snowballed documents
re-enter screening and are ranked on the same terms as searched ones.

\section{Reading and the Evidence Store}
\label{sec:evidence}

Each document in the reading queue is read in its own \code{Send} branch,
against all subquestions assigned to it, in a single call. The reader's
main output is a list of evidence items (\cref{lst:evidence}). A good claim
is a single, self-contained, checkable assertion that names the method, the
baseline and the task. ``Our method outperforms the baseline'' is useless
once separated from the paper.

\begin{lstlisting}[caption={An evidence item (fields abridged).},label={lst:evidence},captionpos=b]
{ "claim":         "<method> improves <metric> on <benchmark> from X% to Y% over <baseline>",
  "support":       "Table 3: ...",            "location": "Section 4.2",
  "evidence_type": "experimental",            "strength": 0.55,
  "subquestion_id":"RQ3",  "doc_id": "doc_0050",  "is_limitation": false }
\end{lstlisting}

\paragraph{Strength is a judgement about evidence, not about assertion.}
The reader is told to score strength low for single-benchmark results
presented as general, missing baselines, missing ablations, unstated
sample sizes, and results visible only in an abstract. In the default
abstract-only mode it must set \code{location} to ``Abstract'' and keep
strength at or below $0.6$, since it cannot verify the experimental setup.
With \code{--full-text}, PDFs are fetched and split into sections, so
locations refer to real sections. Each evidence item is attributed to
exactly one subquestion. A finding that serves two subquestions is emitted
twice, with the claim phrased for each.

\paragraph{Claim clusters and support.} The flat, append-only evidence list
is grouped into \emph{claim clusters} keyed on normalized claim text. A
cluster is \emph{weakly supported} when
\begin{equation}
\bigl|\{\mathrm{doc}(e) : e \in \text{cluster}\}\bigr| < 2
\quad\text{or}\quad
\overline{\mathrm{strength}} < 0.5 .
\end{equation}
This is the distinction between ``the literature shows $X$'' and ``one paper
claimed $X$,'' and the critic acts on it. Clustering on exact text is a
deliberate lower bound. Reconciling semantically equivalent claims needs
judgement, so it is left to the subquestion synthesizer.

\paragraph{Diversity.} A \emph{diversity report} is computed over documents
that actually contributed evidence, not over everything retrieved: a corpus
of 200 papers in which eight supply all the evidence is concentrated. It
records the number of distinct first authors and venues, the share of
evidence supplied by the top document and by the top author group, the
preprint and grey-literature shares, the median year, and the share from
the last three years. It flags \emph{concentration} when the top document
supplies more than 35\% of the evidence, the top author group more than
30\%, or there are fewer than four distinct first authors.

\section{Synthesis, Reflection and the Adaptive Loop}
\label{sec:critic}

\paragraph{Synthesis.} Subquestion synthesis fans out, one branch per
subquestion that has evidence. Each branch combines its evidence into an
answer, states what the evidence does \emph{not} establish, and scores its
own coverage. Global synthesis then reads those answers and builds the
cross-cutting narrative and the report outline. It derives that outline
from the findings rather than repeating the initial subquestion list,
because the right structure for a literature review usually becomes clear
only after the literature has been read.

\paragraph{The critic.} The critic assesses eight dimensions: coverage,
evidence quality, citation coverage, source diversity, contradictions,
recency, knowledge gaps, and search saturation. Its inputs include the
weakly supported claim clusters, the diversity report, and the yield
history of each snowball round and each intent. It does not return a
binary verdict. It returns \emph{research actions}: new search intents,
snowball requests, newly detected contradictions, and plan edits
(\cref{tab:edits}). Knowledge gaps stated as free text are attached to the
subquestion they most likely concern by token overlap. Ties are broken
toward subquestions the critic itself flagged as weak, then toward the one
with the lowest coverage.

\begin{table}[t]
\centering
\small
\caption{Plan edits the critic can issue. Applying them is deterministic.}
\label{tab:edits}
\begin{tabularx}{\linewidth}{@{}lX@{}}
\toprule
Edit & Effect on the subquestion tree \\
\midrule
\code{add}           & New top-level subquestion with origin \code{critic} (default priority $0.7$). \\
\code{split}         & Parent marked \code{superseded} and keeps its evidence; children \code{RQ}$n$\code{.1}, \code{.2}, \ldots\ inherit its priority and receive new searches. \\
\code{rephrase}      & Question text replaced, typically to adopt terminology the literature uses. \\
\code{deprioritize}  & Status \code{deprioritized}, priority capped at $0.2$; excluded from mean coverage. \\
\code{mark\_answered}& Status \code{answered}; no further searches are directed at it. \\
\bottomrule
\end{tabularx}
\end{table}

\paragraph{Termination.} The critic decides whether the evidence is
\emph{sufficient}. The budget decides whether the run may
\emph{continue}, because a critic that keeps asking for one more round
still has to be stopped. The decision follows a fixed order:
\begin{enumerate}[leftmargin=*]
  \item no critique was produced $\Rightarrow$ report (defensive stop);
  \item the critic judges the evidence sufficient $\Rightarrow$ report;
  \item the iteration, reading or search-operation budget is used up
    $\Rightarrow$ report;
  \item the critic asks to continue but proposes no actionable intent,
    snowball request, \code{add} or \code{split} $\Rightarrow$ report,
    because otherwise the loop would spin without changing the state;
  \item otherwise, apply the critique and return to intent generation.
\end{enumerate}
The run records a readable reason for stopping, which is included in the
report.

\section{Report Generation}
\label{sec:report}

The report is written section by section, following the outline produced by
global synthesis. Each section receives only the evidence that belongs to
it. This keeps citations accurate and stops a long report from degrading
toward the end. Documents are cited inline by registry identifier
(\code{[doc\_0012]}). After generation, any identifier not in the registry
is removed from its bracketed group, and the reference list is built from
the documents actually cited in the body. A citation can therefore point
only to a document that was retrieved. Some sections are produced
deterministically from the state rather than by the model: the scope and
method section, the reading guide, the source-diversity caveats,
contradictions, limitations, and the reason the run stopped. If no evidence
was collected, a diagnostic report explains why instead of an empty
synthesis.

The run produces a browsable corpus on disk: the research plan, the full
state as JSON (which \code{--resume} reads), a document index, per-document
metadata with provenance and reading notes, per-subquestion syntheses, the
evidence store grouped by claim cluster, contradictions, and the final
report. The state is checkpointed after every graph step, so an interrupted
run leaves a usable partial corpus.

\section{Implementation}
\label{sec:impl}

\sys{} is about 9{,}500 lines of Python (3.11+) on LangGraph, LangChain's
provider-agnostic \code{init\_chat\_model}, Pydantic and httpx, with
Anthropic, OpenAI, Google and Ollama back-ends available as optional
extras. Configuration has four layers, lowest priority first: code
defaults, a committed \code{optimistresearch.yaml} (merged key by key along the
directory path), \code{.env} for secrets, and environment variables, which
the CLI's flags also set.

\paragraph{Local models.} A fully local configuration needs no API keys.
We found three factors that decide whether a local model works here, and
parameter count matters least. (i) Structured output is constrained to the
JSON schema during decoding, so output always parses and the question is
whether its \emph{content} is good. (ii) Ollama's default context window
is 4{,}096 tokens, and it truncates silently. A screening prompt containing
ten abstracts exceeds that, so the context size is set on every call. (iii)
Mixing models across roles makes Ollama unload and reload weights between
calls, which dominates runtime, so every role should use one model.
Among the models we tested, \code{gemma4} produced critiques grounded in
the actual state. \code{gpt-oss} passed a screening test but judged an
obviously relevant paper irrelevant during extraction and returned no
evidence, a failure that a general benchmark would not have revealed.

\paragraph{Testing.} 284 offline tests cover the deterministic machinery:
the deduplication hierarchy, reducers under fan-in, ranking and allocation,
snowball saturation, source parsing against recorded API responses, and
corpus round-tripping. They also run the whole graph end to end against a
fake model and fake sources. The tests do not measure the quality of model
output. For that, a separate script runs a candidate model against the
real schemas and prompts and flags the failure modes that matter: keeping
a deliberately off-topic paper, rejecting a relevant one, and copying
search intents from the prompt's own examples. Another script checks the
source adapters against the live APIs. Scholarly APIs reject the entire
request when one field is wrong, so a misspelled field name disables a
whole source while every offline test still passes. Three real adapter
bugs were found this way.

\section{Case Study}
\label{sec:case}

We ran \sys{} on the question \emph{``recent research in AI agent memory''}
with the scope note \emph{``focus on empirical evaluations since 2023.''}
Every role used \code{gpt-4o-mini}. The enabled sources were arXiv,
OpenAlex and Crossref, reading was abstract-only, and the budget was 40
documents, 30 search operations, 2 snowball rounds and 3 iterations.
\Cref{tab:case} summarizes the run, and \cref{tab:trace} traces it. About
3.5~minutes passed between the planner's output and the finished report.

\begin{table}[t]
\centering
\small
\begin{minipage}[t]{0.44\linewidth}
\centering
\caption{Run summary.}
\label{tab:case}
\begin{tabular}{@{}lr@{}}
\toprule
Metric & Value \\
\midrule
Research iterations          & 2 \\
Search operations            & 30 / 30 \\
Snowball rounds              & 2 / 2 \\
LLM calls                    & 116 \\
Canonical documents          & 237 \\
\quad via search (depth 0)   & 189 \\
\quad via backward snowball  & 48 \\
Documents read               & 28 / 40 \\
Documents cited in report    & 22 \\
Evidence items               & 87 \\
\quad experimental / survey  & 38 / 24 \\
\quad theoretical / benchmark& 19 / 6 \\
\quad flagged as limitations & 10 \\
Mean evidence strength       & 0.59 \\
Subquestions (final)         & 6 \\
Mean coverage (critic)       & 0.75 \\
\bottomrule
\end{tabular}
\end{minipage}%
\hfill
\begin{minipage}[t]{0.53\linewidth}
\centering
\caption{Condensed trace of the run.}
\label{tab:trace}
\begin{tabular}{@{}p{0.06\linewidth}p{0.88\linewidth}@{}}
\toprule
It & Event \\
\midrule
1 & 5 subquestions planned, 9 intents, routed to 18 queries. \\
  & 195 raw hits $\to$ 129 canonical (66 merged). \\
  & 16 selected for reading (30 doc--SQ pairs). \\
  & Snowball round 1: 4 backward seeds, 48 candidates, 48 new ($\rho=1.0$). \\
  & Snowball round 2: same 4 seeds, 48 candidates, 0 new ($\rho=0$). \\
  & 14 read, 46 evidence items; coverage 0.75--0.85 per SQ. \\
  & Critic: \emph{continue}, coverage 0.80; \code{add} RQ6 (user trust and satisfaction), 4 intents. \\
\midrule
2 & 4 new intents, 12 queries (3 intents closed), 105 hits $\to$ 60 new. \\
  & 15 selected; 14 read, 1 judged not relevant. \\
  & 87 evidence items; RQ3 coverage 0.90. \\
  & Critic: \emph{continue}; proposes \code{mark\_answered} RQ3, \code{deprioritize} RQ4. \\
  & Search budget used up $\Rightarrow$ report. \\
\bottomrule
\end{tabular}
\end{minipage}
\end{table}

\begin{callout}{accent2}{What worked}
The adaptive loop behaved as designed. After the
first iteration, the critic found a dimension the planner had not
anticipated (the effect of memory management on user trust and
satisfaction) and added it as RQ6. RQ6 then received its own intents and
evidence. In the second iteration, one paper (\code{doc\_0139}) that had
already been read for five subquestions was read again only for RQ6, and
three more evidence items were extracted for it. This shows the per-pair
accounting working: earlier document--subquestion pairs were not paid for
again. Reading allocation spread the budget across subquestions, and every
subquestion ended with at least seven evidence items. Canonicalization
merged 34\% of first-iteration hits (66 of 195), and 45 of 105 hits in the
second iteration resolved to documents already in the corpus. The mean
evidence strength of 0.59 sits right at the abstract-only cap, as expected
when no experiment can be checked.
\end{callout}

\begin{callout}{warn}{What the run exposed}
The case study also showed six
weaknesses, which we report because they are informative. The first two
have since been fixed in the code.
\begin{enumerate}[leftmargin=*]
  \item \textbf{Repeated snowball seeds.} The second round chose the
    \emph{same} four seeds in the same direction and spent a round
    fetching 48 edges that were already in the corpus. The seed selector
    excluded snowballed documents as seeds but not documents that had
    already been expanded. The saturation rule needs two low-yield rounds,
    so the round cap ended snowballing, not saturation. \emph{Fixed:} the
    state now records expanded (seed, direction) pairs, and later rounds
    and critic requests skip them.
  \item \textbf{Inflated intent-yield counts.} When RQ6 was added, all 237
    documents were re-screened against it, and each re-screened document
    credited its originating intents again. One intent therefore reports
    18 relevant documents out of 15 found. The critic reads these yields
    when judging saturation. \emph{Fixed:} yields are now recomputed from
    the corpus as distinct documents found by each intent's searches,
    rather than incremented on every screening pass.
  \item \textbf{No arXiv hits.} arXiv returned zero hits for every query
    in both iterations, while OpenAlex and Crossref returned 135 and 60 in
    the first. The literal, AND-combined matching suggests the
    LLM-generated arXiv queries were too specific, but we have not
    confirmed the cause. Either way, a source that returns nothing still
    costs search budget.
  \item \textbf{Weak corroboration signal.} Only 2 of 237 canonical
    documents were returned by more than one source, so the
    multi-source-corroboration term in \cref{eq:rank} was nearly constant.
    With only two sources returning results, this signal adds little.
  \item \textbf{Unused final critique.} The second critique's edits (close
    RQ3, deprioritize RQ4) were recorded but not applied, because the
    search budget stopped the run first. This follows from the termination
    order, but the report should present these edits as recommended next
    steps.
  \item \textbf{Misleading message on failure.} In an earlier run where the
    model provider's API key was not loaded, every LLM call failed and each
    stage fell back to its default. The run completed and produced a
    diagnostic report instead of crashing, but it gave its stopping reason
    as ``the research critic judged the evidence sufficient.'' The fallback
    query also passed the user's raw question, including a trailing
    ``?'', to OpenAlex, which rejected it as an invalid wildcard. Both
    problems are small, but they make diagnosis harder exactly when it is
    needed.
\end{enumerate}
\end{callout}

\section{Discussion and Limitations}
\label{sec:discussion}

\paragraph{Evaluation.} This paper describes a system and does not
evaluate it in a controlled way. We have not yet measured recall against
human-curated reviews, the precision of extracted claims, or the
contribution of each component, such as decomposition versus a single
query or snowballing versus none. The state records what such ablations
need, including per-intent yield, the provenance of every document, and
the snowball depth at which each was found. Running them is the most
important next step.

\paragraph{Abstract-only evidence.} The default reading mode extracts
evidence from abstracts. It is fast and cheap, but it limits how strong any
claim can be, and the evidence store reflects this honestly. Full-text
reading is supported but has not yet been evaluated at scale.

\paragraph{Claim reconciliation.} Exact-text claim clustering
underestimates how many documents support a claim, because paraphrased
claims form separate clusters. A concept graph or embedding-based
clustering layer would fit naturally into the evidence store and would
make the weak-support signal more reliable.

\paragraph{LLM judgement remains the bottleneck.} Deterministic machinery
cannot recover from screening or extraction errors. In our model tests,
the most damaging errors were those that general benchmarks do not
measure: rejecting a clearly relevant paper, and copying examples from the
prompt as search intents. Model choice for the screener and reader roles
should be validated on the system's own schemas.

\section{Conclusion}

\sys{} treats literature research as iterative evidence acquisition over an
explicit, mutable state. In this setting the research plan is an output of
the investigation as well as an input, the result of reading is structured
evidence rather than prose, and everything that does not require judgement
is deterministic code. This design makes the system auditable: every
citation in the report can be traced to a claim, every claim to a document, and
every document to the queries and intents that found it. The same
traceability made the weaknesses in our case study visible, including
repeated snowball seeds, inflated intent yields and silent sources, the
first two of which have since been fixed. We
expect this property to be as useful when improving the system as when
using it.

\bibliographystyle{plainnat}
\bibliography{references}

\end{document}